\documentclass{article} % For LaTeX2e
\usepackage[preprint]{neurips_2025}

\usepackage[T1]{fontenc}
\usepackage{microtype}
\usepackage{hyperref}
\usepackage{url}
\usepackage{booktabs}
\usepackage{lineno}
\usepackage{graphicx}
\usepackage{subcaption}
\usepackage{enumitem}
\usepackage{xspace}
\usepackage{color}
\usepackage[dvipsnames]{xcolor}
\usepackage{multicol}
\usepackage{multirow}
\setitemize{itemsep=-2pt}
\newcommand{\textunknown}{\texttt{Unknown}\xspace}

\definecolor{darkblue}{rgb}{0, 0, 0.5}
\hypersetup{colorlinks=true, citecolor=darkblue, linkcolor=darkblue, urlcolor=darkblue}

\usepackage{listings}

\title{Are vision language models robust to uncertain inputs?}

\author{%
  Xi Wang\\
  Department of Computer Science\\
  Johns Hopkins University\\
  \texttt{xidulu@gmail.com}
  \And
  Eric Nalisnick\\
  Department of Computer Science\\
  Johns Hopkins University\\
  \texttt{nalisnick@jhu.edu}
}

\begin{document}

\maketitle

\begin{abstract}
Robustness against uncertain and ambiguous inputs is a critical challenge for deep learning models.
While recent advancements in large scale vision language models (VLMs, e.g.\ GPT-4o) might suggest that increasing model and training dataset size would mitigate this issue, our empirical evaluation shows a more complicated picture.
Testing models using two classic uncertainty quantification tasks, anomaly detection and classification under inherently ambiguous conditions, we find that newer and larger VLMs indeed exhibit improved robustness compared to earlier models, but still suffer from a tendency to strictly follow instructions, often causing them to hallucinate confident responses even when faced with unclear or anomalous inputs.
Remarkably, for natural images such as ImageNet, this limitation can be overcome without pipeline modifications: simply prompting models to abstain from uncertain predictions enables significant reliability gains, achieving near-perfect robustness in several settings.
However, for domain-specific tasks such as galaxy morphology classification, a lack of specialized knowledge prevents reliable uncertainty estimation.
Finally, we propose a novel mechanism based on caption diversity to reveal a model’s internal uncertainty, enabling practitioners to predict when models will successfully abstain without relying on labeled data.
% \en{no mention to failures (e.g. Galaxy Zoo)?  Abstract says its complicated but then it sounds not so complicated.}
% \xw{Talk about they are equipped with the ability, but just need to activate them through prompting}
\end{abstract}

\section{Introduction}
Uncertainty quantification of neural networks is an important problem widely studied by the deep learning community, especially in real-world, safety-critical settings such as self-driving cars \citep{bojarski2016end} and medical imaging \citep{esteva2017dermatologist}. Traditional vision models often struggle with uncertainty quantification~\citep{nixon2019measuring, guo2017calibration,minderer2021revisiting,gal2016dropout, ovadia2019can}, largely due to their training regime: These models are typically small-scale and trained on specific, highly curated datasets for single tasks.
Consequently, when faced with out-of-distribution (OOD) inputs, they frequently make incorrect yet highly confident predictions, posing significant risks in downstream decision-making systems.

To systematically evaluate the uncertainty quantification ability, two of the most widely adopted workload problems are inputs with corruption and OOD inputs from alternative datasets  (see Sec.~\ref{sec:history} for a more thorough discussion).
Under corrupted inputs, such as those found in CIFAR-10C~\citep{hendrycks2019benchmarking}, models encounter \textbf{covariate shift}, leading to degraded accuracy, in which case a model with good uncertainty quantification should lower its confidence accordingly, rather than maintaining overconfidence despite deteriorating performance, a property knwon as calibration~\citep{guo2017calibration}.
OOD inputs typically refer to inputs from a domain different from the training dataset (e.g. test a model trained on CIFAR-10 with SVHN digits), often having significant \textbf{concept shift} (i.e.\ entirely different label space). The goal of the evaluation is to test  whether a model can identify such inputs via flagging them as uncertain rather than forcing misclassifications.

Vision language models~\citep[VLMs,][]{liu2023llava, gao2023llama, liu2024improved}, such as GPT4o, on the other hand, represent a paradigm shift in visual reasoning. These models leverage massive multimodal datasets and undergo self-supervised pre-training followed by extensive instruction tuning, enabling them to perform diverse tasks in a zero-shot manner, using only the image and a natural language prompt at inference time.
Importantly, the \emph{vast} pretraining corpora makes it unclear if the \textbf{covariate shift} and \textbf{concept shift} challenge underlying the aforementioned two evaluation tasks would still hold for VLM, which raises the question: \emph{Are the traditional evaluation tasks of corruption robustness and OOD detection meaningful for VLMs?}

\begin{figure}
    \centering
    \includegraphics[width=\linewidth]{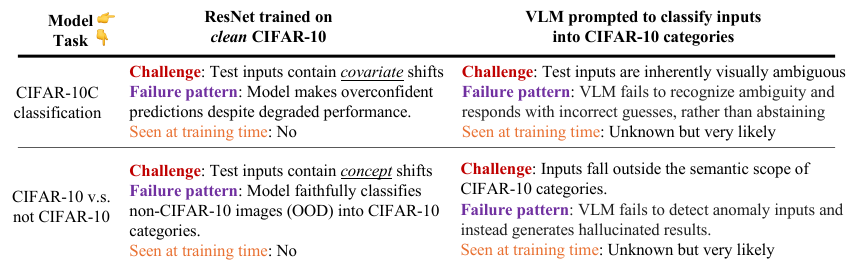}
    \caption{
    \textbf{Classic uncertainty quantification tasks revisited in the VLMs era.}
    Using CIFAR-10 \emph{as an example}, we illustrate how corrupted inputs and inputs from outside CIFAR-10 concepts expose \emph{different} challenges and failure modes in small supervised models vs. large vision language models (VLMs, e.g. GPT4o) prompted to do classification, despite sharing the \emph{same} evaluation data.
    }\vspace{-10pt}
    \label{fig:task_table}
\end{figure}

We argue that these tasks still remain highly relevant to the problem of uncertainty quantification, albeit under a new lens. While VLMs may no longer face clear-cut in-distribution v.s. out-of-distribution boundaries, they still encounter practical challenges in handling \textbf{visually ambiguous inputs}, or \textbf{anomaly inputs} that fall outside the semantic scope defined by a user prompt. These scenarios frequently arise in real-world applications and can reveal significant reliability gaps.
In Fig.~\ref{fig:task_table}, we present illustrative examples of these challenges, adapted for CIFAR-10 domain.

\begin{figure}[t!]
    \centering
    \begin{subfigure}{\linewidth}
    \includegraphics[width=\linewidth]{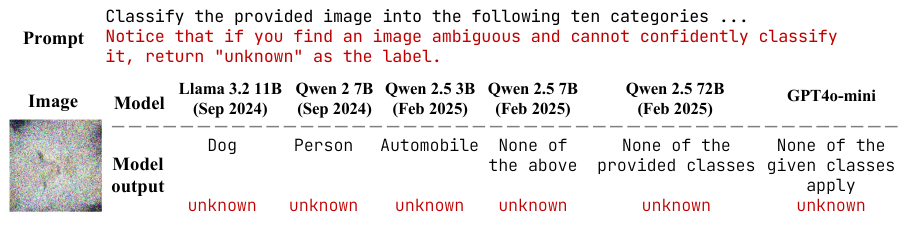}
    \caption{
    VLMs are prompted to classify a noisy cat ImageNet-C image into CIFAR-10 categories. Black \texttt{monospaced text} shows the standard prompt and the corresponding output. {\textcolor{BrickRed}{\texttt{Red}}} text shows the appended explicit rejection prompt and the resulting output. Older and smaller models (e.g.\ Llama and Qwen 2.5 3B) hallucinate labels without the rejection prompt, whereas larger models often reject uncertain inputs even without explicit instructions.
    % We prompted VLMs to classify a noisy cat image from ImageNet-C (left) into CIFAR-10 categories. Black and red \texttt{monospaced text} shows the standard classification prompt and the appended rejection string, and the corresponding outputs. Older and smaller VLMs would make random guesses (Llama, Qwen 2.5 3B) or hallucinate a class not provided (Qwen 2) without the rejection string. Newer and larger VLMs can automatically ``reject'' the inputs \emph{without} any special instruction.
    }
    \label{fig:no_special_prompt_result}
    \end{subfigure}
    
    \begin{subfigure}{\linewidth}
        \centering
        \includegraphics[width=\linewidth]{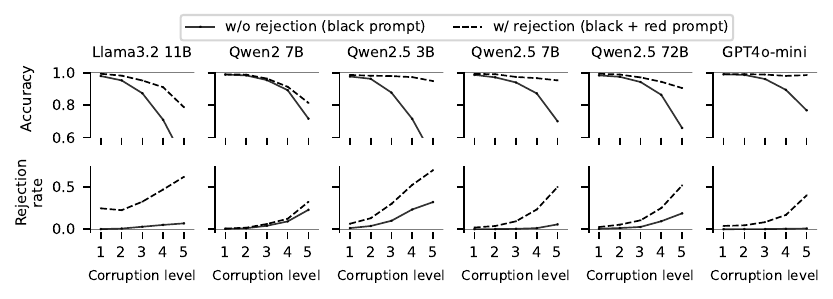}
        \caption{
        We prompt different VLMs (columns) to classify the same 1,000 samples subset of Gaussian noise-corrupted ImageNet into CIFAR-10 categories, under various corruption intensities (x-axis). Under standard prompt (black part in top of \ref{fig:no_special_prompt_result} only), the accuracy (among all samples the models output a proper label) decreases as corruption intensifies (solid lines).
        When models are explicitly prompted to reject ambiguous inputs (black and red parts), the accuracy for classified samples becomes significantly higher (dashed lines). The bottom row shows the corresponding rejection rates for each model: The models reject more inputs as corruption levels increase.
        }\label{fig:corrputed_acc}
    \end{subfigure}
    \caption{
    \textbf{VLMs show degraded performance under corrupted inputs, allowing rejection helps maintain reliability.}
    \emph{Top row} demonstrates VLMs' outputs for a selected sample with or without a rejection prompt appended.
    \emph{Bottom row} shows the classification accuracy under standard prompt without (solid line) vs. with (dashed line) rejection instruction prompt appended.
    % However, the overall performance under corruption without special prompts drops significantly (solid line, Fig.~\ref{fig:corrputed_acc}), explicitly rejection prompting enables the model to abstain more samples, hence achieving better accuracy (dashed line, Fig.~\ref{fig:corrputed_acc}). 
    % \xw{Verbally describe the rejection string here???}
    % \xw{Mention that here accuracy means the correct ratio among all inputs that the models give a proper label}
    }
    \label{fig:corrupted_results}
\end{figure}

Therefore, in this paper, we empirically studied VLMs behavior on these two problems (Sec.~\ref{sec:eval_task}). 
We begin by evaluating VLM performance on corrupted ImageNet images, finding that although out-of-the-box performance reveals notable vulnerabilities, larger and newer VLMs exhibit improved robustness (Fig.~\ref{fig:no_special_prompt_result}). Furthermore, we show that prompting the models explicitly to "reject ambiguous inputs" substantially enhances their reliability, enabling the models to abstain from making predictions when appropriate (Fig.~\ref{fig:corrputed_acc}).
Additionally, we study a classic anomaly detection setting using CIFAR-10 vs. non-CIFAR-10 images, where the goal is to determine whether models can correctly reject inputs that fall outside the specified label space, where we again find that simple prompting is sufficient for models to identify and reject anomalies.
Lastly, building on these findings, we extend our analysis to more domain-specific and specialized tasks, such as ECG signal classification and galaxy image recognition, where we find that without sufficient domain expertise, VLMs may show degraded or complete failure at providing reliable uncertainty quantification.
% \xw{Add argument on why galaxy zoo needs domain knowledge}

Finally, we propose a method to reveal VLM's uncertain level about an input image (Sec.~\ref{sec:caption_diversity}), by prompting the VLM to generate multiple captions for the input image under random sampling decoding, we find that VLMs tend to generate more diverse captions for visually ambiguous samples, which is more likely to be abstained, and vice versa.
This insight allows us to predict the model's ability to successfully perform classification with rejection without relying on labeled ground truth.

\section{Background and related work}
\vspace{-5pt}
% \xw{TODO: Make the terminology cleaner and more consistent, maybe refer to Pixmix paper's related work section for some ideas how to phrase things}

\subsection{Improving models' robustness against uncertain inputs}
\label{sec:history}

\begin{figure}[ht]
    \centering
    \includegraphics[width=\linewidth]{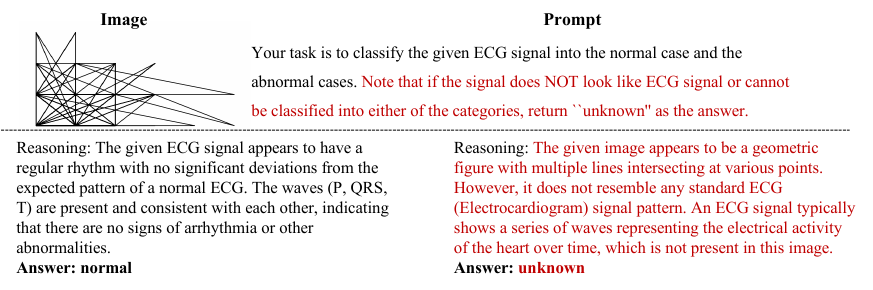}
    \caption{\textbf{Enabling the rejection option allows VLM to pick out anomaly inputs, preventing hallucination.} When prompted to classify a random line image into normal v.s. abnormal ECG signal, certain VLM (Qwen 2.5 3B) would generate hallucinated results (left), potentially caused by its tendency towards strictly following instructions~\citep{sharma2023towards}. 
    However, when the prompt explicitly permits rejecting non-ECG input (additional {\color{BrickRed}{red}} texts in the prompt), the same model correctly identifies the input as anomalous and responds with \textunknown (right). 
    }
    \label{fig:ecg_example}
\end{figure}

Neural networks often struggle with input uncertainty due to their training methodology. When trained on carefully curated datasets with minimal ambiguity, models develop a tendency to produce high-confidence predictions for all inputs, regardless of their clarity or relevance. This behavior creates significant reliability concerns when deploying these models in real-world scenarios where inputs may be ambiguous, corrupted, or entirely outside the scope of the model's expertise. To systematically evaluate models' behavior under uncertain inputs, the deep learning community has identified two primary testing paradigms:

\begin{itemize}[leftmargin=*, topsep=0pt, parsep=0pt, itemsep=1.5pt]
\item \textbf{Distribution / covariate shift} occurs when a model encounters inputs from a different distribution than its training data, while maintaining the same label space. Examples include synthetic corruptions or stylistic variations~\citep[ImageNet-C/P,][]{hendrycks2019benchmarking}, or alternative data collection pipelines~\citep[ImageNet-V2,][]{recht2019imagenet}. In these scenarios, models often show degraded accuracy, as these inputs are outside their capacity, while maintaining high confidence predictions. Such a dangerous combination will provide misleading signals for downstream decision-making, whereas ideally, a model should elicit some signals indicating its ``unsureness'' and express confidence proportional to its likelihood of correctness~\citep{guo2017calibration}.
\item \textbf{Concept shift (OOD/anomaly detection)} typically considers anomaly inputs coming from entirely different datasets with non-overlapping label spaces than the training distribution. For instance, a digit classifier might encounter images of animals or vehicles. In such cases, the desired behavior is for the model to recognize the mismatch and refuse classification entirely, rather than confidently assigning inputs to irrelevant categories. This capability, often called OOD/anomaly detection, is crucial for safe deployment in open-world environments~\citep{nalisnick2018deep,nalisnick2019detecting}.
\end{itemize}

Standard deep neural networks without specialized training typically fail at both tasks. Researchers have developed numerous approaches to address these limitations. The central idea of many of these methods lies around tweaking the training loop in a way such that the model becomes capable of eliciting uncertainty information through the statistics (e.g.\ entropy) of the predictive distribution, such as Bayesian neural networks~\citep{mackay1992practical,neal2012bayesian,graves2011practical,blundell2015weight,gal2016dropout,maddox2019simple,aitchison2020bayesian,aitchison2020statistical,daxberger2021laplace, daxberger2021bayesian, nalisnick2019dropout,  izmailov2021bayesian, wenzel2020good}, deep ensemble~\citep{lakshminarayanan2017simple, abe2022deep,  d'angelo2021repulsive}, outlier exposure~\citep{hendrycks2018deep}, with which practitioners can be aware of when distribution shift / anomaly inputs shows up.
There are also methods aiming at improving the models' generalization to certain distribution shift, such as test time adaptation~\citep{wang2020tent,wang2023robustness, schirmer2024test,nado2020evaluating, schneider2020improving} for improving accuracy against corrupted inputs, and to anomaly inputs, such as open set classification~\citep{geng2020recent} or meta learning \citep{hospedales2021meta}.

% To our knowledge, no prior research has evaluated VLMs on these two evaluation settings \xw{maybe change the claim / phrasing}.
Initially, one might think that such tasks pose \emph{little} challenge for VLMs, based on two arguments:
\begin{enumerate}[leftmargin=*, topsep=0pt, parsep=0pt, itemsep=1.5pt]
\item \textbf{No clear OOD boundary}: Traditional models (e.g. ResNet trained on CIFAR-10) face clear distribution boundaries; in contrast, VLMs trained on extensive and diverse multimodal datasets inherently blur these distinctions.
\item \textbf{Scaling improves uncertainty quantification}: Prior observations suggest that \emph{simply} scaling models and datasets can significantly mitigate uncertainty quantification problems. Indeed, prior studies highlight improvements in OOD detection~\citep{fort2021exploring} and calibration under distribution shift~\citep{minderer2021revisiting} with larger-scale models.
\end{enumerate}

However, upon closer investigation, these arguments do not fully hold:
\begin{enumerate}[leftmargin=*, topsep=0pt, parsep=0pt, itemsep=1.5pt]
\item Indeed the distribution shift challenges represented by some of the dataset, such as ImageNet-V2/P may no longer be applicable for VLMs, there still exists certain real-world uncertainties, such as \emph{corruptions caused by sensor malfunctions} or \emph{accidental user uploads}, that can be simulated by datasets like ImageNet-C~\footnote{\small{Corrupted inputs are typically seen as a distribution shift from clean training distribution, but the ``noise inherent in the observations''~\citep[definition of aleatoric uncertainty,][]{kendall2017uncertainties} also introduces ambiguous visual cues, making the corrupted inputs hard to recognize visually.}}
and \emph{anomaly detection settings}, that remain practically significant.
% These inputs can lead to serious reliability issues if not adequately addressed.
\item State-of-the-art \emph{large-scale} VLMs may exhibit non-negligible issues due to their instruction-following nature~\citep{sharma2023towards}, when no special prompt is included, they frequently resort to incorrect or random predictions when faced with uncertain or anomalous inputs
\end{enumerate}

Importantly, conventional methods for improving robustness, such as Bayesian neural networks or ensemble techniques, are impractical for VLMs due to the enormous computational overhead required for fine-tuning. Similarly, test-time interventions like test time adaptation or temperature scaling are difficult to implement given the black-box nature of many of these models.

Encouragingly, we discovered that the implicit uncertainty quantification capabilities embedded in VLMs, derived from their instruction-following behavior, offer a simple yet effective solution. Directly prompting VLMs to reject uncertain inputs rather than forcing classification significantly enhances their robustness without necessitating architectural changes or specialized training. This implicit form of uncertainty handling capitalizes on the model's inherent ability to condition its behavior on input instructions, thus providing practitioners with a straightforward approach to improving reliability in practical deployments.

\vspace{-5pt}
\subsection{Uncertainty quantification in large language models}
\vspace{-5pt}

A large body of work investigates uncertainty quantification in large language models (LLMs), in the absence of multimodal inputs. One line of research explores the confidence score elicted by the model itself, where models are prompted to express their own certainty about their outputs~\citep{kadavath2022language}. Another class of methods leverages sampling-based approaches, such as measuring the number of semantic clusters among independently sampled completions to infer semantic consistency and model confidence~\citep{kuhn2023semantic, nikitin2024kernel}. 
We hypothesize that such techniques are transferable to multimodal tasks, including those we examine. However, as shown in our experiments, prompting the model with an additional line of instruction that enables rejection is sufficient for the task we considered. Nevertheless, the aforementioned approaches could augment this with more explicit and a numerically-scored uncertainty estimates.

% Particularly relevant is the work by~\citet{hou2023decomposing}, which decomposes uncertainty into epistemic and aleatoric components in language models. They find that aleatoric uncertainty that arises from inherent ambiguity in language can be captured through ensembling over rephrased queries. This aligns with our observation that caption diversity in visual inputs reflects ambiguity: when the model perceives the image as ambiguous, it generates a broader set of plausible captions. In both cases, diversity in generated outputs can serve as a proxy for input uncertainty, be it textual or visual.
% \vspace{-10pt}\
\vspace{-5pt}
\subsection{Uncertainty quantification in vision language models}
\vspace{-5pt}
% \vspace{-10pt}

Several recent works have explored uncertainty quantification in the context of vision-language \emph{embedding models} such as CLIP~\citep{radford2021learning} and DINO~\citep{caron2021emerging}, focusing on the quality and uncertainty of the learned representations~\citep{miao2024bayesian, fillioux2024foundation, cui2024exploring}. However, these embedding models are not the focus of our work, we are studying LlaVA~\citep{liu2023llava} style full VLM that can follow natural language instructions.
% Vision-language models (VLMs) typically consist of a vision-language embedding component followed by a pure large language model (LLM), as in models like LLaVA~\citep{liu2023llava}. Thus, it is unclear how uncertainty estimation techniques developed for embeddings can be directly applied to full VLMs that generate natural language responses.

In the domain of full VLMs, a few studies examine their uncertainty estimation capabilities. For instance, \citet{kostumov2024uncertainty} and \citet{groot2024overconfidence} investigate the calibration of VLMs on \emph{standard} Visual Question Answering (VQA) benchmarks, paralleling the calibration protocols established in the LLM literature discussed in the previous section. However, these works focus on standard benchmarks where there is limited inherent uncertainty in the inputs. Recent work \citet{miyai2024generalized, miyai2024unsolvable} introduce the concept of Unsolvable Problem Detection (UPD), where the goal is to determine whether a given image-question pair is unanswerable. Their Intrinsically Visual Question Detection (IVQD) task is most similar to our anomaly detection setup, in which a mismatch between the question and image indicates an uncertain or ambiguous case. However, our evaluation differs in construction: while they rely on manually annotated datasets, we automatically generate our test data in a manner aligned with classic out-of-distribution (OOD) detection literatures, enabling more scalable evaluation. Moreover, their work focuses on uncertainty arising from the interplay between visual and textual modalities, our experiments additionally studies the ambiguity and uncertainty from images alone. 
\citet{zhang2024vl} considers the interaction of VLMs with corrupted image, similar to our work, however their goal is to perform \emph{hallucination detection} by checking VLMs output variability under various corrupted versions of the same input, while our work aims at evaluating VLMs' \emph{robustness} against corrupted inputs.
% This distinction opens a complementary research avenue, where the source of uncertainty—visual, textual, or combinatorial—can be explicitly categorized and targeted.

\vspace{-5pt}
\section{Methods}
\vspace{-5pt}

\subsection{Evaluation of VLM's implicit uncertainty quantification ability}
\label{sec:eval_task}

\begin{figure}[!t]
     \centering
    \begin{subfigure}[t]{.47\linewidth}
    \includegraphics[width=\linewidth]{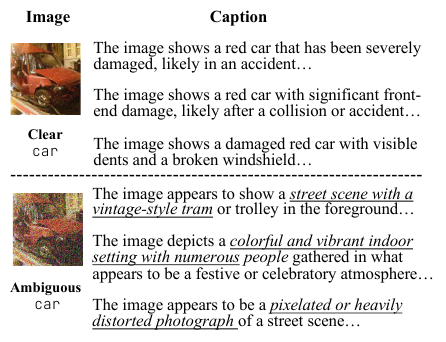}
    \caption{ImageNet-C}
    \label{fig:imgnet_capt}
    \end{subfigure}
    \begin{subfigure}[t]{.495\linewidth}
    \includegraphics[width=\linewidth]{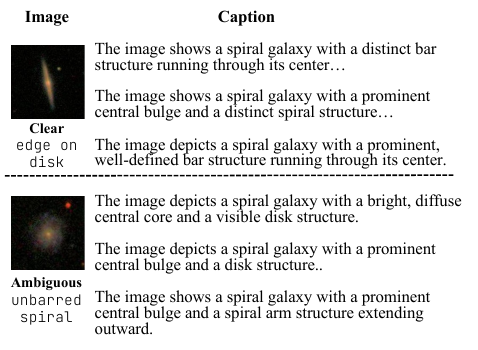}
    \caption{Galaxy Zoo}
    \label{fig:gz2_cap}
    \end{subfigure}
    \caption{
    \textbf{VLMs generate diverse captions for ambiguous images.}
    We prompt Qwen2.5 7B to ``generate a description'' given an input image under different random seeds. Clean image from ImageNet receives consistent captions while its corrupted version having a diverse set of captions (top left v.s. bottom left).
    For the galaxy image where annotators show significant disagreement on whether there exist spiral arms (bottom right), VLMs fail to have diversity in the caption, indicating that the model does not understand the ambiguity, likely due to limited domain knowledge.
    % Three independently randomly generated captions. Bottom right shows an Galaxy image, where annotators show significant disagreement on whether there there exists spiral arms (\texttt{smooth round} v.s. \texttt{unbarred spiral})
    }\vspace{-8pt}
    \label{fig:caption_viz}
\end{figure}

Our main contribution centers on assessing VLMs' ability to recognize and express uncertainty through \emph{rejecting} problematic inputs, when performing image classification tasks through natural language prompting. Drawing inspiration from the literature on out-of-distribution robustness (Sec.~\ref{sec:history}), we design two complementary evaluation tasks:

\textbf{Anomaly detection with prompting}~ This task evaluates a model's ability to identify inputs that fall outside the provided category definitions. We prompt the model to reject inputs that do not belong to any of the specified categories. For \emph{evaluation}, we treat samples inside the given categories as the negative class and other samples as the positive class, employing standard binary classification metrics: precision (what fraction of rejected inputs are actual anomalies) and recall (what fraction of all anomalies are successfully rejected), to measure performance.

\textbf{Classification with rejection against ambiguous inputs with prompting}~ In this task, we present models with inputs that exhibit varying degrees of inherent ambiguity, making them potentially classifiable into multiple categories. We then explicitly prompt the model to reject inputs it finds difficult to classify into a single category. Our hypothesis is that as ambiguity increases, models without a rejection option will resort to random guessing, leading to increased error rates. Conversely, models with a rejection option can abstain from classification rather than make low-confidence predictions. To \emph{evaluate} performance, we measure accuracy on the subset of non-rejected samples, with the ideal behavior being perfect accuracy after rejection, demonstrating the model's ability to recognize when it might make errors.

While these evaluation paradigms root in the study of out-of-distribution robustness, a concept not applicable to VLMs, they still represent real-world challenges that VLMs must overcome in practical applications. Additionally, unlike previous approaches that required model modifications or specialized training, our method for uncertainty quantification leverages the inherent capabilities of VLMs through just prompting. It is also worth noting that our method for quantification uncertainty is \emph{implicit} in that our model only has a binary option: Classify the inputs or reject, instead of eliciting a continuous score for uncertainty level.

% Lastly, the tasks we consider align closely with the concept of aleatoric uncertainty ~\citep{smith2024rethinking} that stems from inherent data ambiguity rather than model limitations, and thus cannot be reduced through additional data collection.

\vspace{-5pt}
\subsection{Caption diversity for understanding the underlying mechanism of rejection}
\label{sec:caption_diversity}
\vspace{-5pt}

When adopting uncertainty quantification methods for non-black-box models, we typically have access to a continuous score that reflects \emph{how uncertain the model finds a given input to be}. Common examples include the degree of disagreement among ensemble components~\citep{abe2022deep}, the typicality of a test input relative to the training distribution~\citep{nalisnick2019detecting}, or the distance of test inputs from the training data measured in a kernel space~\citep{liu2020simple, immer2021improving}.
In contrast, when VLMs are prompted to reject ambiguous inputs, the only available feedback is binary: whether the input was classified or rejected.
This raises a critical question, can we have a ``uncertainty score'' that tells us \emph{how uncertain an image is for a particular VLM}, such that the higher score an input receives, the more likely it will be rejected by a VLM.

We propose using caption diversity as a reflection for VLM's uncertainty in an input. Intuitively, when a model encounters an ambiguous image, one permitting multiple plausible interpretations, it produces a more diverse set of captions across independent generations (under random sampling decoding). Conversely, clear and unambiguous images yield consistent descriptions.
To quantify this, we compute a \textbf{caption diversity score} by first embedding all generated captions with a sentence transformer model~\citep[\texttt{all-mpnet-base-v2} from][]{reimers-2019-sentence-bert}, then calculating one minus the averaged pairwise cosine similarity among the embeddings. 

\begin{figure}[!t]
    \centering
    \includegraphics[width=.97\linewidth]{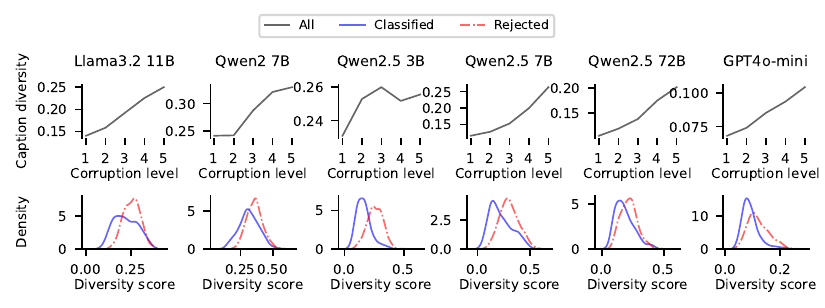}\vspace{-5pt}
    \caption{
     \textbf{Caption diversity reflects model uncertainty under ambiguous inputs.}
     We empirically verified the hypothesis from Sec.~\ref{sec:caption_diversity}.
    \emph{Top:} As corruption increases, caption diversity rises across all models, indicating greater uncertainty.
    \emph{Bottom:} {\color{Red}{Rejected samples}} exhibit higher caption diversity than {\color{blue}{classified ones}}, suggesting that diversity of independently-generated captions correlates with models' internal uncertainty level for an input and the tendency for rejecting it when prompted.
    }\vspace{-5pt}
    \label{fig:cap_diversity}
\end{figure}

Our experiments indeed support this hypothesis (Fig.~\ref{fig:imgnet_capt} and Fig.~\ref{fig:cap_diversity}): As input images become more ambiguous due to higher corruption level, the overall caption diversity increases. Additionally, the images model chooses to reject consistently exhibits higher caption diversity than the classified ones.

Beyond serving as an analytical tool, caption diversity also offers a practical mechanism for predicting rejection behavior without labeled data. By examining the relationship between diversity scores and input ambiguity, practitioners can assess whether a model is likely to abstain from unreliable predictions. Notably, our experiments reveal that for specialized domains requiring expert knowledge, such as galaxy morphology, models generate low-diversity captions even for ambiguous inputs, failing to recognize their own uncertainty (Fig.~\ref{fig:gz2_cap}) and leading to ineffective rejection (Fig.~\ref{fig:galaxy_results}).

% \begin{figure}[!t]
%     \centering
%     \includegraphics[height=1.22cm]{figs/model_legend_gz2.pdf}\includegraphics[height=1.2cm]{figs/prompt_legend_new.pdf}\\[-1.2ex]
%     \centering
%     \includegraphics[width=.95\linewidth]{figs/gz2_results_new.pdf}
%     \caption{Classification with rejection fails does NOT work on Galaxy Zoo. \xw{TODO: Finish the caption}}
%     \label{fig:enter-label}
% \end{figure}

\vspace{-5pt}
\section{Experiments}
\label{sec:experiments}
\vspace{-5pt}

In this section, we empirically evaluate the two tasks proposed in Sec.~\ref{sec:eval_task} on the following families of VLMs: GPT4o-mini~\citep{achiam2023gpt}, Llama 3.2~\citep{dubey2024llama}, and Qwen 2/2.5 ~\citep{wang2024qwen2, bai2025qwen2}. GPT4o-mini has a version of \texttt{2024-07-18}, Llama 3.2 is released in September 2024, Qwen 2 and 2.5 are released in September 2024 and February 2025 respectively. For the Qwen2.5 72B, we use the official released AWQ quantized version.

For all experiments, we treat all models as black boxes, where we compute the metrics by looking at the output string without using the information from the logits. Unless explicitly stated, we use deterministic sampling for querying the VLMs. For experiments that use multiple generations from random decoding, we use a temperature of 0.6, a top-P of 0.95, and top-K of 50 for all models. We conduct all experiments on an internal cluster of Nvidia H100s and A100s.

The complete prompt used are shown in Appendix.~\ref{appendix:full_prompt}. We conduct ablation study over the prompting style (Appendix.~\ref{appendix:abl_study}): Broadly, we find that the ability of rejection is sensitive to prompting style for Llama and Qwen2, but the rest models are less sensitive to the prompting mode.

\vspace{-5pt}
\subsection{Anomaly detection}
\vspace{-5pt}

\begin{table*}[t!]
\footnotesize
\centering
\begin{tabular}{lcccccc}
\toprule
\multirow{2}{*}{Model} &
  \multicolumn{3}{c}{CIFAR-10 v.s. Not CIFAR-10} &
  \multicolumn{3}{c}{ECG v.s. Not ECG} \\
& \textbf{Precision} $\uparrow$ & \textbf{Recall} $\uparrow$ & \textbf{F1} $\uparrow$
& \textbf{Precision} $\uparrow$ & \textbf{Recall} $\uparrow$ & \textbf{F1} $\uparrow$ \\
\hline
Llama 3.2 11B         & 0.991 & 0.718 & 0.833 & 0.698 & 0.308 & 0.426 \\
Qwen 2 7B           & 0.964 & 0.757 & 0.848 & 0.998 & 0.994 & 0.996 \\
Qwen 2.5 3B         & 0.993 & 0.782 & 0.875 &  0.598 & 1.000 & 0.749 \\ 
Qwen 2.5 7B    & 0.982 & 0.967 & 0.974 & 0.907 & 0.972 & 0.939 \\
Qwen 2.5 72B      & 0.976 & 0.986 & 0.981 & 0.398   & 1.000   & 0.570 \\
GPT4o-mini        & 0.964 & 0.974 & 0.969 & 0.360 & 1.000 & 0.529 \\
\end{tabular}
\caption{
\textbf{VLMs can perform anomaly detection}. Results are evaluated using precision, recall, and F1-score across two anomaly detection tasks where anomaly  inputs are considered positive cases. In both tasks, VLMs achieve high recall, successfully identifying most anomalous inputs. However, for the ECG task, models exhibit lower overall performance (lower F1), primarily due to low precision caused by over-rejection by frequently abstaining even on valid inputs.
}\vspace{-10pt}
\label{tab:ood_detection}
\end{table*}

We begin by examining anomaly detection tasks across two datasets:

\textbf{CIFAR-10 v.s. Not CIFAR-10}: We selected images from ImageNet~\citep{deng2009imagenet} with concepts overlapping with CIFAR-10 categories as the target classification samples (detailed class mapping shown in Appendix.~\ref{appendix:cifar_imgnet_mapping}), then considered images outside these concepts as anomalies. We prompted the models to perform classification on target samples, i.e.\ those classifiable into CIFAR-10 categories, and to identify and reject the anomalous samples outside CIFAR-10 concepts. This setup is known to be challenging for OOD/anomaly detection methods~\citep{yang2023imagenet} as the anomalous samples are visually very similar to classifiable samples, since they come from the same dataset. We construct the evaluation dataset with 3,000 images in total, composed of 1,800 classifiable images and 1,200 anomaly ones, where a random classifier would give a precision of 0.4 and a recall of 0.5.

\textbf{ECG v.s. Not ECG}: We used ECG signals from the PTB database~\citep{wagner2020ptb}, treat the signals as images, prompting the VLM to classify signals as normal or abnormal, as well as identifying and rejecting anomaly inputs, which we construct with randomly generated line patterns (example shown in the top left panel of Figure 2, detailed in Appendix.~\ref{appendix:anomaly_construction}). Our evaluation dataset consists of 500 images each from the normal, abnormal, and anomaly categories, therefore, a random guess baseline should have a precision and recall value of $1/3$

Fig.~\ref{fig:ecg_example} illustrates a representative example from the ECG vs. not ECG experiment. Without explicit instructions to reject anomaly inputs, certain VLMs (e.g., Qwen 2.5 3B) hallucinate answers when presented with random lines, attempting to force classification despite the input clearly not being an ECG. However, when the prompt is augmented with a single line allowing rejection of non-ECG inputs (highlighted in red in the prompt), the same model correctly identifies the input as anomalous and responds with \textunknown.

The quantitative results are shown in Table.~\ref{tab:ood_detection}, where we evaluate the performance using precision, recall, and F1 value as discussed in Sec.~\ref{sec:eval_task}.
For the CIFAR-10 vs. Not CIFAR-10 task, all models achieve high precision larger than 0.96, indicating that models are not showing ``over refusal'', but with varying recall rates, indicating some models (e.g. Qwen2.5 v.s. Qwen 2) can more reliably identify anomalous inputs when instructed to do so than others. For ECG vs. Not ECG task, overall performance drops significantly as indicated by the lower F1 value.
Llama 3.2 11B shows limited ability to identify anomalous ECG inputs (0.308 recall), while Qwen 2 and Qwen 2.5 models demonstrate remarkable precision and recall. Interestingly, GPT4o-mini shows perfect recall but lower precision (0.360), suggesting a tendency toward over-rejecting valid ECG signals as anomalous.

\vspace{-5pt}
\subsection{Classification with rejection under ambiguous inputs}
\vspace{-5pt}

\begin{figure}[!t]
    \centering
    \includegraphics[width=.34\linewidth]{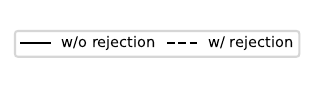}\\[-2.8ex]
    \centering
    \includegraphics[width=.49\linewidth]{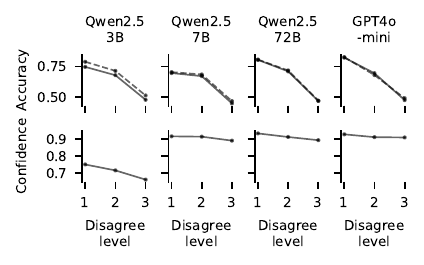}
    \includegraphics[width=.498\linewidth]{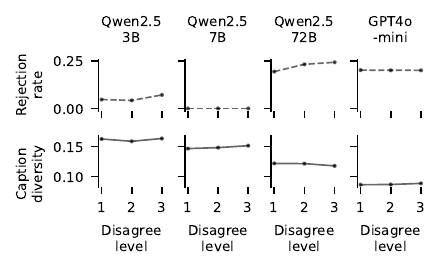}\vspace{-5pt}
    \caption{
     \textbf{Classification with rejection fails on Galaxy Zoo.}
As annotator disagreement level increases (higher input ambiguity), VLMs' accuracy degrades, but do not effectively utilize rejection to improve performance (top left). Caption diversity and confidence remain \emph{flat} (bottom row), indicating that models fail to recognize the uncertainty in this domain, likely due to insufficient domain knowledge.
}\vspace{-10pt}
    \label{fig:galaxy_results}
\end{figure}

For classification with rejection, we consider the following two problems: Classifying corrupted ImageNet images into CIFAR-10 categories and morphological classification of Galaxy Zoo images. The reason behind choosing these two datasets is that they have ground truth ambiguity levels provided, such that we could better understand how VLM behaves as the level of uncertainty varies.

\textbf{Classify ImageNet-C into CIFAR-10 categories} Similar to the anomaly detection setting, we again selected some classes from ImageNet that overlaps with CIFAR-10 categories (detailed in Appendix.~\ref{appendix:cifar_imgnet_mapping}), then we used their corrupted version to introduce ambiguity in the input, we considered 4 types of corruption: Gaussian noise, defocus blur, pixelated, all from ImageNet-C~\citep{hendrycks2019benchmarking}, and pixmix from \citet{hendrycks2022pixmix}. For all corruption types and levels, we use the same 1,000 randomly selected images.

\textbf{Classification of Galaxy Zoo images} We considered Galaxy Zoo 2, a crowdsourced dataset for galaxy morphological classification. Importantly, each image comes with an annotator agreement score (\texttt{leaf\_prob} in the metadata table), where a low agreement score implies that an image is ambiguous among classes due to, e.g.\ unclear visual features. We select a 5,000 sample subset from the dataset and prompt the model to classify them into the four categories provided from Galaxy MNIST~\citep{galaxy_mnist}. We present more details on dataset construction in Appendix.~\ref{appendix:gz2_explain}.

We begin by looking at the VLM's behavior on images corrupted by Gaussian noise, as shown in Fig.~\ref{fig:corrupted_results}.
Before explicit prompting VLMs to reject ambiguous samples, we first look at VLM's behavior when prompted naively just to classify the image, where we find that some models can already reject highly noised inputs (Fig.~\ref{fig:no_special_prompt_result}) instead of hallucinating an answer, however if we look at their accuracy, we can see that it still drops significantly as corruption level increases (solid line, Fig.~\ref{fig:corrputed_acc}).
Now when they are \emph{explicitly} prompted to reject ambiguous samples, the accuracy among classified samples becomes much higher, indicating that the models are internally aware of the uncertainty in the input, but it needs to be ``activated'' via prompting.
We observe similar results for other corruption types shown in Appendix.~\ref{appendix:all_corruption}.
Additionally, we studied caption diversity (Sec.~\ref{sec:caption_diversity}), for each image, we prompt the model to independently generate 20 captions. Broadly, we find that as the corruption level increases, the overall caption diversity increases (Fig.~\ref{fig:cap_diversity} top); additionally, rejected samples are those that receive higher caption diversity (Fig.~\ref{fig:cap_diversity} bottom, red v.s. blue lines), indicating that indeed caption diversity reflects VLMs' internal uncertainty level on input images, aligning with our hypothesis.

However, this is not the case for Galaxy Zoo, the results are shown in Fig.~\ref{fig:galaxy_results}, where we see that the accuracy significantly drops as ambiguity level increases, but rejection option only provides marginal improvement, indicating that the VLMs do not understand the ambiguity, such result is also predicted if we look at the caption diversity, which stays constant as annotator diagreement level increases, in contrast to the pattern for ImageNet-C images, an illustrative comparison is provided in Fig.~\ref{fig:caption_viz}.
Notice that we also study just the output confidence, i.e. we prompt the model to randomly generate multiple answers and look at the maximum softmax probability of the averaged prediction vector, and again, we did not see any sign that the model is aware of the ambiguity.
We hypothesize that such task would require more domain specific knowledge with customized image embedding model~\citep{walmsley2022towards, parker2024astroclip}, indeed as verified by \citet{wang2024effective}, a general purpose CLIP model significantly \emph{underperforms} fine-tuned CLIP on galaxy classification task, however building a VLM with domain knowledge goes beyond the scope of this work.

\vspace{-5pt}
\section{Conclusion}\vspace{-5pt}
To summarize, our work revisits two classic uncertainty quantification evaluation settings: handling corrupted inputs and anomaly detection, which are challenging for small-scale models trained from scratch due to the inputs' OOD nature. While the boundary of OOD may be less clear for VLMs, we argue that these tasks still represent real-world challenges involving \emph{inherent} data ambiguity and anomalousness, issues that cannot be resolved through model scaling alone.
We evaluate VLMs on these two problems and find that, for standard benchmarks (e.g.\, ImageNet-C and CIFAR-10 vs. Not CIFAR-10), models generally perform well through providing explicit rejection option in prompt. 

However, in problems requiring specialized expert knowledge, such as galaxy classification, VLMs consistently exhibit suboptimal performance, which highlights the importance of domain-specific foundation models: These models are not only essential for achieving strong task performance but are also critical for ensuring reliability. Without a proper understanding of the input domain, models struggle to recognize and quantify uncertainty, since a good understanding of the input could be a necessary prerequisite for understanding the uncertainty associated with it.

\section{Limitation}
\label{sec:dis_and_lim}

A key limitation of the current work is the lack of a solution for handling failure cases in classification with rejection on the Galaxy Zoo dataset (Fig. 6). This issue likely stems from the model's insufficient domain knowledge of galaxy morphology. One potential direction for addressing this challenge is to augment VLMs with expert knowledge, such as galaxy decision trees~\citep{walmsley2022galaxy}, which explicitly guide the model to extract relevant features from galaxy images in a step-by-step manner.

% In this paper, we revisit two good old BNN tasks, and we argue that they still represent real world challenges. We evaluate VLMs performance on them, where we find that: 1. To perform uncertainty quantification, now we can only do prompting; 2. As the model's capacity increases, 

% Overall, they do not pose too much challenges 

% This can be roughly understood as, the model can only understand aleatoric uncertainty if epistemic uncertainty is low / reduced. However, it is still unclear how to detect images with high epistemic uncertainty. In particular, we believe this is still an open challenge for LLMs in general: How do we know when we can trust a model's behavior in a certain task or not?

% Also our proposed method does not provide explicit uncertainty quantification, i.e. one that provides you with a numerical score, therefore you cannot manually set a threshold for rejection. 

% Lastly, we did not have a solution for the galaxy image case, this highlights the importance of having VLM with domain expertise: If you want to reason about uncertainty, a pre-requisite is for the model to understand the task, since uncertainty quantification is a task more difficult.

\newpage
\bibliography{iclr2024_conference}

\begin{thebibliography}{70}
\providecommand{\natexlab}[1]{#1}
\providecommand{\url}[1]{\texttt{#1}}
\expandafter\ifx\csname urlstyle\endcsname\relax
  \providecommand{\doi}[1]{doi: #1}\else
  \providecommand{\doi}{doi: \begingroup \urlstyle{rm}\Url}\fi

\bibitem[Abe et~al.(2022)Abe, Buchanan, Pleiss, Zemel, and Cunningham]{abe2022deep}
Taiga Abe, Estefany~Kelly Buchanan, Geoff Pleiss, Richard Zemel, and John~P Cunningham.
\newblock Deep ensembles work, but are they necessary?
\newblock \emph{Advances in Neural Information Processing Systems}, 35:\penalty0 33646--33660, 2022.

\bibitem[Achiam et~al.(2023)Achiam, Adler, Agarwal, Ahmad, Akkaya, Aleman, Almeida, Altenschmidt, Altman, Anadkat, et~al.]{achiam2023gpt}
Josh Achiam, Steven Adler, Sandhini Agarwal, Lama Ahmad, Ilge Akkaya, Florencia~Leoni Aleman, Diogo Almeida, Janko Altenschmidt, Sam Altman, Shyamal Anadkat, et~al.
\newblock Gpt-4 technical report.
\newblock \emph{arXiv preprint arXiv:2303.08774}, 2023.

\bibitem[Aitchison(2020{\natexlab{a}})]{aitchison2020bayesian}
Laurence Aitchison.
\newblock Bayesian filtering unifies adaptive and non-adaptive neural network optimization methods.
\newblock \emph{Advances in Neural Information Processing Systems}, 33:\penalty0 18173--18182, 2020{\natexlab{a}}.

\bibitem[Aitchison(2020{\natexlab{b}})]{aitchison2020statistical}
Laurence Aitchison.
\newblock A statistical theory of cold posteriors in deep neural networks.
\newblock \emph{arXiv preprint arXiv:2008.05912}, 2020{\natexlab{b}}.

\bibitem[Bai et~al.(2025)Bai, Chen, Liu, Wang, Ge, Song, Dang, Wang, Wang, Tang, et~al.]{bai2025qwen2}
Shuai Bai, Keqin Chen, Xuejing Liu, Jialin Wang, Wenbin Ge, Sibo Song, Kai Dang, Peng Wang, Shijie Wang, Jun Tang, et~al.
\newblock Qwen2. 5-vl technical report.
\newblock \emph{arXiv preprint arXiv:2502.13923}, 2025.

\bibitem[Blundell et~al.(2015)Blundell, Cornebise, Kavukcuoglu, and Wierstra]{blundell2015weight}
Charles Blundell, Julien Cornebise, Koray Kavukcuoglu, and Daan Wierstra.
\newblock Weight uncertainty in neural network.
\newblock In \emph{International conference on machine learning}, pp.\  1613--1622. PMLR, 2015.

\bibitem[Bojarski et~al.(2016)Bojarski, Del~Testa, Dworakowski, Firner, Flepp, Goyal, Jackel, Monfort, Muller, Zhang, et~al.]{bojarski2016end}
Mariusz Bojarski, Davide Del~Testa, Daniel Dworakowski, Bernhard Firner, Beat Flepp, Prasoon Goyal, Lawrence~D Jackel, Mathew Monfort, Urs Muller, Jiakai Zhang, et~al.
\newblock End to end learning for self-driving cars.
\newblock \emph{arXiv preprint arXiv:1604.07316}, 2016.

\bibitem[Caron et~al.(2021)Caron, Touvron, Misra, J{\'e}gou, Mairal, Bojanowski, and Joulin]{caron2021emerging}
Mathilde Caron, Hugo Touvron, Ishan Misra, Herv{\'e} J{\'e}gou, Julien Mairal, Piotr Bojanowski, and Armand Joulin.
\newblock Emerging properties in self-supervised vision transformers.
\newblock In \emph{Proceedings of the IEEE/CVF international conference on computer vision}, pp.\  9650--9660, 2021.

\bibitem[Cui et~al.(2024)Cui, He, Zhang, Deng, Dong, and Zhu]{cui2024exploring}
Peng Cui, Guande He, Dan Zhang, Zhijie Deng, Yinpeng Dong, and Jun Zhu.
\newblock Exploring aleatoric uncertainty in object detection via vision foundation models.
\newblock \emph{arXiv preprint arXiv:2411.17767}, 2024.

\bibitem[D'Angelo \& Fortuin(2021)D'Angelo and Fortuin]{d'angelo2021repulsive}
Francesco D'Angelo and Vincent Fortuin.
\newblock Repulsive deep ensembles are bayesian.
\newblock In A.~Beygelzimer, Y.~Dauphin, P.~Liang, and J.~Wortman Vaughan (eds.), \emph{Advances in Neural Information Processing Systems}, 2021.
\newblock URL \url{https://openreview.net/forum?id=LAKplpLMbP8}.

\bibitem[Daxberger et~al.(2021{\natexlab{a}})Daxberger, Kristiadi, Immer, Eschenhagen, Bauer, and Hennig]{daxberger2021laplace}
Erik Daxberger, Agustinus Kristiadi, Alexander Immer, Runa Eschenhagen, Matthias Bauer, and Philipp Hennig.
\newblock Laplace redux-effortless bayesian deep learning.
\newblock \emph{Advances in Neural Information Processing Systems}, 34, 2021{\natexlab{a}}.

\bibitem[Daxberger et~al.(2021{\natexlab{b}})Daxberger, Nalisnick, Allingham, Antor{\'a}n, and Hern{\'a}ndez-Lobato]{daxberger2021bayesian}
Erik Daxberger, Eric Nalisnick, James~U Allingham, Javier Antor{\'a}n, and Jos{\'e}~Miguel Hern{\'a}ndez-Lobato.
\newblock Bayesian deep learning via subnetwork inference.
\newblock In \emph{International Conference on Machine Learning}, pp.\  2510--2521. PMLR, 2021{\natexlab{b}}.

\bibitem[Deng et~al.(2009)Deng, Dong, Socher, Li, Li, and Fei-Fei]{deng2009imagenet}
Jia Deng, Wei Dong, Richard Socher, Li-Jia Li, Kai Li, and Li~Fei-Fei.
\newblock Imagenet: A large-scale hierarchical image database.
\newblock In \emph{2009 IEEE conference on computer vision and pattern recognition}, pp.\  248--255. Ieee, 2009.

\bibitem[Dubey et~al.(2024)Dubey, Jauhri, Pandey, Kadian, Al-Dahle, Letman, Mathur, Schelten, Yang, Fan, et~al.]{dubey2024llama}
Abhimanyu Dubey, Abhinav Jauhri, Abhinav Pandey, Abhishek Kadian, Ahmad Al-Dahle, Aiesha Letman, Akhil Mathur, Alan Schelten, Amy Yang, Angela Fan, et~al.
\newblock The llama 3 herd of models.
\newblock \emph{arXiv preprint arXiv:2407.21783}, 2024.

\bibitem[Esteva et~al.(2017)Esteva, Kuprel, Novoa, Ko, Swetter, Blau, and Thrun]{esteva2017dermatologist}
Andre Esteva, Brett Kuprel, Roberto~A Novoa, Justin Ko, Susan~M Swetter, Helen~M Blau, and Sebastian Thrun.
\newblock Dermatologist-level classification of skin cancer with deep neural networks.
\newblock \emph{nature}, 542\penalty0 (7639):\penalty0 115--118, 2017.

\bibitem[Fillioux et~al.(2024)Fillioux, Silva-Rodr{\'\i}guez, Ayed, Courn{\`e}de, Vakalopoulou, Christodoulidis, and Dolz]{fillioux2024foundation}
Leo Fillioux, Julio Silva-Rodr{\'\i}guez, Ismail~Ben Ayed, Paul-Henry Courn{\`e}de, Maria Vakalopoulou, Stergios Christodoulidis, and Jose Dolz.
\newblock Are foundation models for computer vision good conformal predictors?
\newblock \emph{arXiv preprint arXiv:2412.06082}, 2024.

\bibitem[Fort et~al.(2021)Fort, Ren, and Lakshminarayanan]{fort2021exploring}
Stanislav Fort, Jie Ren, and Balaji Lakshminarayanan.
\newblock Exploring the limits of out-of-distribution detection.
\newblock \emph{Advances in neural information processing systems}, 34:\penalty0 7068--7081, 2021.

\bibitem[Gal \& Ghahramani(2016)Gal and Ghahramani]{gal2016dropout}
Yarin Gal and Zoubin Ghahramani.
\newblock Dropout as a bayesian approximation: Representing model uncertainty in deep learning.
\newblock In \emph{international conference on machine learning}, pp.\  1050--1059. PMLR, 2016.

\bibitem[Gao et~al.(2023)Gao, Han, Zhang, Lin, Geng, Zhou, Zhang, Lu, He, Yue, et~al.]{gao2023llama}
Peng Gao, Jiaming Han, Renrui Zhang, Ziyi Lin, Shijie Geng, Aojun Zhou, Wei Zhang, Pan Lu, Conghui He, Xiangyu Yue, et~al.
\newblock Llama-adapter v2: Parameter-efficient visual instruction model.
\newblock \emph{arXiv preprint arXiv:2304.15010}, 2023.

\bibitem[Geng et~al.(2020)Geng, Huang, and Chen]{geng2020recent}
Chuanxing Geng, Sheng-jun Huang, and Songcan Chen.
\newblock Recent advances in open set recognition: A survey.
\newblock \emph{IEEE transactions on pattern analysis and machine intelligence}, 43\penalty0 (10):\penalty0 3614--3631, 2020.

\bibitem[Graves(2011)]{graves2011practical}
Alex Graves.
\newblock Practical variational inference for neural networks.
\newblock \emph{Advances in neural information processing systems}, 24, 2011.

\bibitem[Groot \& Valdenegro-Toro(2024)Groot and Valdenegro-Toro]{groot2024overconfidence}
Tobias Groot and Matias Valdenegro-Toro.
\newblock Overconfidence is key: Verbalized uncertainty evaluation in large language and vision-language models.
\newblock \emph{arXiv preprint arXiv:2405.02917}, 2024.

\bibitem[Guo et~al.(2017)Guo, Pleiss, Sun, and Weinberger]{guo2017calibration}
Chuan Guo, Geoff Pleiss, Yu~Sun, and Kilian~Q Weinberger.
\newblock On calibration of modern neural networks.
\newblock In \emph{International conference on machine learning}, pp.\  1321--1330. PMLR, 2017.

\bibitem[Hendrycks \& Dietterich(2019)Hendrycks and Dietterich]{hendrycks2019benchmarking}
Dan Hendrycks and Thomas Dietterich.
\newblock Benchmarking neural network robustness to common corruptions and perturbations.
\newblock \emph{arXiv preprint arXiv:1903.12261}, 2019.

\bibitem[Hendrycks et~al.(2018)Hendrycks, Mazeika, and Dietterich]{hendrycks2018deep}
Dan Hendrycks, Mantas Mazeika, and Thomas Dietterich.
\newblock Deep anomaly detection with outlier exposure.
\newblock \emph{arXiv preprint arXiv:1812.04606}, 2018.

\bibitem[Hendrycks et~al.(2022)Hendrycks, Zou, Mazeika, Tang, Li, Song, and Steinhardt]{hendrycks2022pixmix}
Dan Hendrycks, Andy Zou, Mantas Mazeika, Leonard Tang, Bo~Li, Dawn Song, and Jacob Steinhardt.
\newblock Pixmix: Dreamlike pictures comprehensively improve safety measures.
\newblock In \emph{Proceedings of the IEEE/CVF Conference on Computer Vision and Pattern Recognition}, pp.\  16783--16792, 2022.

\bibitem[Hospedales et~al.(2021)Hospedales, Antoniou, Micaelli, and Storkey]{hospedales2021meta}
Timothy Hospedales, Antreas Antoniou, Paul Micaelli, and Amos Storkey.
\newblock Meta-learning in neural networks: A survey.
\newblock \emph{IEEE transactions on pattern analysis and machine intelligence}, 44\penalty0 (9):\penalty0 5149--5169, 2021.

\bibitem[Immer et~al.(2021)Immer, Korzepa, and Bauer]{immer2021improving}
Alexander Immer, Maciej Korzepa, and Matthias Bauer.
\newblock Improving predictions of bayesian neural nets via local linearization.
\newblock In \emph{International conference on artificial intelligence and statistics}, pp.\  703--711. PMLR, 2021.

\bibitem[Izmailov et~al.(2021)Izmailov, Vikram, Hoffman, and Wilson]{izmailov2021bayesian}
Pavel Izmailov, Sharad Vikram, Matthew~D Hoffman, and Andrew Gordon~Gordon Wilson.
\newblock What are bayesian neural network posteriors really like?
\newblock In \emph{International Conference on Machine Learning}, pp.\  4629--4640. PMLR, 2021.

\bibitem[Kadavath et~al.(2022)Kadavath, Conerly, Askell, Henighan, Drain, Perez, Schiefer, Hatfield-Dodds, DasSarma, Tran-Johnson, et~al.]{kadavath2022language}
Saurav Kadavath, Tom Conerly, Amanda Askell, Tom Henighan, Dawn Drain, Ethan Perez, Nicholas Schiefer, Zac Hatfield-Dodds, Nova DasSarma, Eli Tran-Johnson, et~al.
\newblock Language models (mostly) know what they know.
\newblock \emph{arXiv preprint arXiv:2207.05221}, 2022.

\bibitem[Kendall \& Gal(2017)Kendall and Gal]{kendall2017uncertainties}
Alex Kendall and Yarin Gal.
\newblock What uncertainties do we need in bayesian deep learning for computer vision?
\newblock \emph{Advances in neural information processing systems}, 30, 2017.

\bibitem[Kostumov et~al.(2024)Kostumov, Nutfullin, Pilipenko, and Ilyushin]{kostumov2024uncertainty}
Vasily Kostumov, Bulat Nutfullin, Oleg Pilipenko, and Eugene Ilyushin.
\newblock Uncertainty-aware evaluation for vision-language models.
\newblock \emph{arXiv preprint arXiv:2402.14418}, 2024.

\bibitem[Kuhn et~al.(2023)Kuhn, Gal, and Farquhar]{kuhn2023semantic}
Lorenz Kuhn, Yarin Gal, and Sebastian Farquhar.
\newblock Semantic uncertainty: Linguistic invariances for uncertainty estimation in natural language generation.
\newblock \emph{arXiv preprint arXiv:2302.09664}, 2023.

\bibitem[Lakshminarayanan et~al.(2017)Lakshminarayanan, Pritzel, and Blundell]{lakshminarayanan2017simple}
Balaji Lakshminarayanan, Alexander Pritzel, and Charles Blundell.
\newblock Simple and scalable predictive uncertainty estimation using deep ensembles.
\newblock \emph{Advances in neural information processing systems}, 30, 2017.

\bibitem[Liu et~al.(2023)Liu, Li, Wu, and Lee]{liu2023llava}
Haotian Liu, Chunyuan Li, Qingyang Wu, and Yong~Jae Lee.
\newblock Visual instruction tuning.
\newblock In \emph{NeurIPS}, 2023.

\bibitem[Liu et~al.(2024)Liu, Li, Li, and Lee]{liu2024improved}
Haotian Liu, Chunyuan Li, Yuheng Li, and Yong~Jae Lee.
\newblock Improved baselines with visual instruction tuning.
\newblock In \emph{Proceedings of the IEEE/CVF Conference on Computer Vision and Pattern Recognition}, pp.\  26296--26306, 2024.

\bibitem[Liu et~al.(2020)Liu, Lin, Padhy, Tran, Bedrax~Weiss, and Lakshminarayanan]{liu2020simple}
Jeremiah Liu, Zi~Lin, Shreyas Padhy, Dustin Tran, Tania Bedrax~Weiss, and Balaji Lakshminarayanan.
\newblock Simple and principled uncertainty estimation with deterministic deep learning via distance awareness.
\newblock \emph{Advances in neural information processing systems}, 33:\penalty0 7498--7512, 2020.

\bibitem[MacKay(1992)]{mackay1992practical}
David~JC MacKay.
\newblock A practical bayesian framework for backpropagation networks.
\newblock \emph{Neural computation}, 4\penalty0 (3):\penalty0 448--472, 1992.

\bibitem[Maddox et~al.(2019)Maddox, Izmailov, Garipov, Vetrov, and Wilson]{maddox2019simple}
Wesley~J Maddox, Pavel Izmailov, Timur Garipov, Dmitry~P Vetrov, and Andrew~Gordon Wilson.
\newblock A simple baseline for bayesian uncertainty in deep learning.
\newblock \emph{Advances in Neural Information Processing Systems}, 32, 2019.

\bibitem[Miao et~al.(2024)Miao, Lei, Zhou, and Deng]{miao2024bayesian}
Yibo Miao, Yu~Lei, Feng Zhou, and Zhijie Deng.
\newblock Bayesian exploration of pre-trained models for low-shot image classification.
\newblock In \emph{Proceedings of the IEEE/CVF Conference on Computer Vision and Pattern Recognition}, pp.\  23849--23859, 2024.

\bibitem[Minderer et~al.(2021)Minderer, Djolonga, Romijnders, Hubis, Zhai, Houlsby, Tran, and Lucic]{minderer2021revisiting}
Matthias Minderer, Josip Djolonga, Rob Romijnders, Frances Hubis, Xiaohua Zhai, Neil Houlsby, Dustin Tran, and Mario Lucic.
\newblock Revisiting the calibration of modern neural networks.
\newblock \emph{Advances in Neural Information Processing Systems}, 34:\penalty0 15682--15694, 2021.

\bibitem[Miyai et~al.(2024{\natexlab{a}})Miyai, Yang, Zhang, Ming, Lin, Yu, Irie, Joty, Li, Li, et~al.]{miyai2024generalized}
Atsuyuki Miyai, Jingkang Yang, Jingyang Zhang, Yifei Ming, Yueqian Lin, Qing Yu, Go~Irie, Shafiq Joty, Yixuan Li, Hai Li, et~al.
\newblock Generalized out-of-distribution detection and beyond in vision language model era: A survey.
\newblock \emph{arXiv preprint arXiv:2407.21794}, 2024{\natexlab{a}}.

\bibitem[Miyai et~al.(2024{\natexlab{b}})Miyai, Yang, Zhang, Ming, Yu, Irie, Li, Li, Liu, and Aizawa]{miyai2024unsolvable}
Atsuyuki Miyai, Jingkang Yang, Jingyang Zhang, Yifei Ming, Qing Yu, Go~Irie, Yixuan Li, Hai Li, Ziwei Liu, and Kiyoharu Aizawa.
\newblock Unsolvable problem detection: Evaluating trustworthiness of vision language models.
\newblock \emph{arXiv preprint arXiv:2403.20331}, 2024{\natexlab{b}}.

\bibitem[Nado et~al.(2020)Nado, Padhy, Sculley, D'Amour, Lakshminarayanan, and Snoek]{nado2020evaluating}
Zachary Nado, Shreyas Padhy, D~Sculley, Alexander D'Amour, Balaji Lakshminarayanan, and Jasper Snoek.
\newblock Evaluating prediction-time batch normalization for robustness under covariate shift.
\newblock \emph{arXiv preprint arXiv:2006.10963}, 2020.

\bibitem[Nalisnick et~al.(2018)Nalisnick, Matsukawa, Teh, Gorur, and Lakshminarayanan]{nalisnick2018deep}
Eric Nalisnick, Akihiro Matsukawa, Yee~Whye Teh, Dilan Gorur, and Balaji Lakshminarayanan.
\newblock Do deep generative models know what they don't know?
\newblock \emph{arXiv preprint arXiv:1810.09136}, 2018.

\bibitem[Nalisnick et~al.(2019{\natexlab{a}})Nalisnick, Hern{\'a}ndez-Lobato, and Smyth]{nalisnick2019dropout}
Eric Nalisnick, Jos{\'e}~Miguel Hern{\'a}ndez-Lobato, and Padhraic Smyth.
\newblock Dropout as a structured shrinkage prior.
\newblock In \emph{International Conference on Machine Learning}, pp.\  4712--4722. PMLR, 2019{\natexlab{a}}.

\bibitem[Nalisnick et~al.(2019{\natexlab{b}})Nalisnick, Matsukawa, Teh, and Lakshminarayanan]{nalisnick2019detecting}
Eric Nalisnick, Akihiro Matsukawa, Yee~Whye Teh, and Balaji Lakshminarayanan.
\newblock Detecting out-of-distribution inputs to deep generative models using typicality.
\newblock \emph{arXiv preprint arXiv:1906.02994}, 2019{\natexlab{b}}.

\bibitem[Neal(2012)]{neal2012bayesian}
Radford~M Neal.
\newblock \emph{Bayesian learning for neural networks}, volume 118.
\newblock Springer Science \& Business Media, 2012.

\bibitem[Nikitin et~al.(2024)Nikitin, Kossen, Gal, and Marttinen]{nikitin2024kernel}
Alexander Nikitin, Jannik Kossen, Yarin Gal, and Pekka Marttinen.
\newblock Kernel language entropy: Fine-grained uncertainty quantification for llms from semantic similarities.
\newblock \emph{Advances in Neural Information Processing Systems}, 37:\penalty0 8901--8929, 2024.

\bibitem[Nixon et~al.(2019)Nixon, Dusenberry, Zhang, Jerfel, and Tran]{nixon2019measuring}
Jeremy Nixon, Michael~W Dusenberry, Linchuan Zhang, Ghassen Jerfel, and Dustin Tran.
\newblock Measuring calibration in deep learning.
\newblock In \emph{CVPR Workshops}, 2019.

\bibitem[Ovadia et~al.(2019)Ovadia, Fertig, Ren, Nado, Sculley, Nowozin, Dillon, Lakshminarayanan, and Snoek]{ovadia2019can}
Yaniv Ovadia, Emily Fertig, Jie Ren, Zachary Nado, David Sculley, Sebastian Nowozin, Joshua Dillon, Balaji Lakshminarayanan, and Jasper Snoek.
\newblock Can you trust your model's uncertainty? evaluating predictive uncertainty under dataset shift.
\newblock \emph{Advances in neural information processing systems}, 32, 2019.

\bibitem[Parker et~al.(2024)Parker, Lanusse, Golkar, Sarra, Cranmer, Bietti, Eickenberg, Krawezik, McCabe, Morel, et~al.]{parker2024astroclip}
Liam Parker, Francois Lanusse, Siavash Golkar, Leopoldo Sarra, Miles Cranmer, Alberto Bietti, Michael Eickenberg, Geraud Krawezik, Michael McCabe, Rudy Morel, et~al.
\newblock Astroclip: a cross-modal foundation model for galaxies.
\newblock \emph{Monthly Notices of the Royal Astronomical Society}, 531\penalty0 (4):\penalty0 4990--5011, 2024.

\bibitem[Radford et~al.(2021)Radford, Kim, Hallacy, Ramesh, Goh, Agarwal, Sastry, Askell, Mishkin, Clark, et~al.]{radford2021learning}
Alec Radford, Jong~Wook Kim, Chris Hallacy, Aditya Ramesh, Gabriel Goh, Sandhini Agarwal, Girish Sastry, Amanda Askell, Pamela Mishkin, Jack Clark, et~al.
\newblock Learning transferable visual models from natural language supervision.
\newblock In \emph{International conference on machine learning}, pp.\  8748--8763. PmLR, 2021.

\bibitem[Recht et~al.(2019)Recht, Roelofs, Schmidt, and Shankar]{recht2019imagenet}
Benjamin Recht, Rebecca Roelofs, Ludwig Schmidt, and Vaishaal Shankar.
\newblock Do imagenet classifiers generalize to imagenet?
\newblock In \emph{International Conference on Machine Learning}, pp.\  5389--5400. PMLR, 2019.

\bibitem[Reimers \& Gurevych(2019)Reimers and Gurevych]{reimers-2019-sentence-bert}
Nils Reimers and Iryna Gurevych.
\newblock Sentence-bert: Sentence embeddings using siamese bert-networks.
\newblock In \emph{Proceedings of the 2019 Conference on Empirical Methods in Natural Language Processing}. Association for Computational Linguistics, 11 2019.
\newblock URL \url{https://arxiv.org/abs/1908.10084}.

\bibitem[Schirmer et~al.(2024)Schirmer, Zhang, and Nalisnick]{schirmer2024test}
Mona Schirmer, Dan Zhang, and Eric Nalisnick.
\newblock Test-time adaptation with state-space models.
\newblock In \emph{ICML 2024 Workshop on Structured Probabilistic Inference $\{$$\backslash$\&$\}$ Generative Modeling}, 2024.

\bibitem[Schneider et~al.(2020)Schneider, Rusak, Eck, Bringmann, Brendel, and Bethge]{schneider2020improving}
Steffen Schneider, Evgenia Rusak, Luisa Eck, Oliver Bringmann, Wieland Brendel, and Matthias Bethge.
\newblock Improving robustness against common corruptions by covariate shift adaptation.
\newblock \emph{Advances in Neural Information Processing Systems}, 33:\penalty0 11539--11551, 2020.

\bibitem[Sharma et~al.(2023)Sharma, Tong, Korbak, Duvenaud, Askell, Bowman, Cheng, Durmus, Hatfield-Dodds, Johnston, et~al.]{sharma2023towards}
Mrinank Sharma, Meg Tong, Tomasz Korbak, David Duvenaud, Amanda Askell, Samuel~R Bowman, Newton Cheng, Esin Durmus, Zac Hatfield-Dodds, Scott~R Johnston, et~al.
\newblock Towards understanding sycophancy in language models.
\newblock \emph{arXiv preprint arXiv:2310.13548}, 2023.

\bibitem[Wagner et~al.(2020)Wagner, Strodthoff, Bousseljot, Kreiseler, Lunze, Samek, and Schaeffter]{wagner2020ptb}
Patrick Wagner, Nils Strodthoff, Ralf-Dieter Bousseljot, Dieter Kreiseler, Fatima~I Lunze, Wojciech Samek, and Tobias Schaeffter.
\newblock Ptb-xl, a large publicly available electrocardiography dataset.
\newblock \emph{Scientific data}, 7\penalty0 (1):\penalty0 1--15, 2020.

\bibitem[Walmsley(2022)]{galaxy_mnist}
Martin Walmsley.
\newblock Galaxymnist: Galaxy images labelled by morphology (shape).
\newblock \url{https://github.com/mwalmsley/galaxy_mnist}, 2022.
\newblock Accessed: 2025-05-14.

\bibitem[Walmsley et~al.(2022{\natexlab{a}})Walmsley, Lintott, G{\'e}ron, Kruk, Krawczyk, Willett, Bamford, Kelvin, Fortson, Gal, et~al.]{walmsley2022galaxy}
Mike Walmsley, Chris Lintott, Tobias G{\'e}ron, Sandor Kruk, Coleman Krawczyk, Kyle~W Willett, Steven Bamford, Lee~S Kelvin, Lucy Fortson, Yarin Gal, et~al.
\newblock Galaxy zoo decals: Detailed visual morphology measurements from volunteers and deep learning for 314 000 galaxies.
\newblock \emph{Monthly Notices of the Royal Astronomical Society}, 509\penalty0 (3):\penalty0 3966--3988, 2022{\natexlab{a}}.

\bibitem[Walmsley et~al.(2022{\natexlab{b}})Walmsley, Slijepcevic, Bowles, and Scaife]{walmsley2022towards}
Mike Walmsley, Inigo~Val Slijepcevic, Micah Bowles, and Anna~MM Scaife.
\newblock Towards galaxy foundation models with hybrid contrastive learning.
\newblock \emph{arXiv preprint arXiv:2206.11927}, 2022{\natexlab{b}}.

\bibitem[Wang et~al.(2020)Wang, Shelhamer, Liu, Olshausen, and Darrell]{wang2020tent}
Dequan Wang, Evan Shelhamer, Shaoteng Liu, Bruno Olshausen, and Trevor Darrell.
\newblock Tent: Fully test-time adaptation by entropy minimization.
\newblock In \emph{International Conference on Learning Representations}, 2020.

\bibitem[Wang et~al.(2024{\natexlab{a}})Wang, Bai, Tan, Wang, Fan, Bai, Chen, Liu, Wang, Ge, et~al.]{wang2024qwen2}
Peng Wang, Shuai Bai, Sinan Tan, Shijie Wang, Zhihao Fan, Jinze Bai, Keqin Chen, Xuejing Liu, Jialin Wang, Wenbin Ge, et~al.
\newblock Qwen2-vl: Enhancing vision-language model's perception of the world at any resolution.
\newblock \emph{arXiv preprint arXiv:2409.12191}, 2024{\natexlab{a}}.

\bibitem[Wang et~al.(2024{\natexlab{b}})Wang, Wang, and Luo]{wang2024effective}
Ruoqi Wang, Haitao Wang, and Qiong Luo.
\newblock Effective fine-tuning of vision-language models for accurate galaxy morphology analysis.
\newblock \emph{arXiv preprint arXiv:2411.19475}, 2024{\natexlab{b}}.

\bibitem[Wang \& Aitchison(2023)Wang and Aitchison]{wang2023robustness}
Xi~Wang and Laurence Aitchison.
\newblock Robustness to corruption in pre-trained bayesian neural networks.
\newblock In \emph{The Eleventh International Conference on Learning Representations}, 2023.
\newblock URL \url{https://openreview.net/forum?id=kUI41mY8bHl}.

\bibitem[Wei et~al.(2022)Wei, Wang, Schuurmans, Bosma, Xia, Chi, Le, Zhou, et~al.]{wei2022chain}
Jason Wei, Xuezhi Wang, Dale Schuurmans, Maarten Bosma, Fei Xia, Ed~Chi, Quoc~V Le, Denny Zhou, et~al.
\newblock Chain-of-thought prompting elicits reasoning in large language models.
\newblock \emph{Advances in neural information processing systems}, 35:\penalty0 24824--24837, 2022.

\bibitem[Wenzel et~al.(2020)Wenzel, Roth, Veeling, {\'S}wi{\k{a}}tkowski, Tran, Mandt, Snoek, Salimans, Jenatton, and Nowozin]{wenzel2020good}
Florian Wenzel, Kevin Roth, Bastiaan~S Veeling, Jakub {\'S}wi{\k{a}}tkowski, Linh Tran, Stephan Mandt, Jasper Snoek, Tim Salimans, Rodolphe Jenatton, and Sebastian Nowozin.
\newblock How good is the bayes posterior in deep neural networks really?
\newblock \emph{arXiv preprint arXiv:2002.02405}, 2020.

\bibitem[Yang et~al.(2023)Yang, Zhang, and Russakovsky]{yang2023imagenet}
William Yang, Byron Zhang, and Olga Russakovsky.
\newblock Imagenet-ood: Deciphering modern out-of-distribution detection algorithms.
\newblock \emph{arXiv preprint arXiv:2310.01755}, 2023.

\bibitem[Zhang et~al.(2024)Zhang, Zhang, and Zheng]{zhang2024vl}
Ruiyang Zhang, Hu~Zhang, and Zhedong Zheng.
\newblock Vl-uncertainty: Detecting hallucination in large vision-language model via uncertainty estimation.
\newblock \emph{arXiv preprint arXiv:2411.11919}, 2024.

\end{thebibliography}
\bibliographystyle{colm2025_conference}

\newpage
\appendix
% \section{Appendix}

\section{Prompt}\label{appendix:full_prompt}
\subsection{Anomaly detection}
\textbf{CIFAR-10 v.s. not CIFAR-10}
\begin{lstlisting}
You are given an image, check if the image is from one of the following 10 classes:
airplane, automobile, bird, cat, deer, dog, frog, horse, ship, and truck, if so, classify the image into one of 10 classes, if not, label them as "unknown".

Think step by step, and give an answer in the following format:
Reasoning: <your_reasoning>
Answer: <your answer>
\end{lstlisting}

\textbf{ECG v.s. not ECG}
\begin{lstlisting}
Your task is to classify the given ECG signal into the normal case and the abnormal cases.

Note that if the signal does NOT look like ECG signal or cannot be classified into
either of the categories, return ``unknown'' as the answer.

Give an answer and step-by-step reasoning in the format:
Reasoning: <your_reasoning>
Answer: normal/abnormal/unknown
\end{lstlisting}

\subsection{Classification with rejection under ambiguous inputs}
\textbf{ImageNet-C}
\begin{lstlisting}
You are given an image, the image belongs to one of the following 10 classes: 
airplane, automobile, bird, cat, deer, dog, frog, horse, ship, and truck.

Please assign each an image a label from the 10 classes, think step by step, and give an answer in the following format:
Reasoning: your_reasoning
Label: class_name

Notice that if you find an image very ambiguous and cannot confidently classify it, return "unknown" as the label.
\end{lstlisting}

\textbf{Galaxy Zoo classification}
\begin{lstlisting}
You are an expert astronomer specializing in galaxy morphology. You will be shown images of galaxies and need to classify them into one of the following categories by analyzing their visual characteristics:

1. smooth_round: Galaxies that appear completely or nearly circular, with a smooth decrease in brightness from center to edge.
2. smooth_cigar: Galaxies that appear elongated and smooth, with an elliptical shape resembling a cigar, showing a gradual decrease in brightness from center to edge.
3. edge_on_disk: Galaxies viewed from the side, appearing as a thin line or disk with a bright central bulge, similar to viewing a dinner plate from its edge.
4. unbarred_spiral: Spiral galaxies without a central bar structure, showing distinct spiral arms emanating directly from the galactic center.

Classify the galaxy into exactly one of the four categories listed above with step by step reasoning.
Notice that if you find an image very ambiguous and cannot confidently classify it,
return "unknown" as the label.

Your response should be structured as:
Reasoning: [Brief explanation of the key visual features that support this classification]
Answer: [category_name] or unknown
\end{lstlisting}

\subsection{Caption generation}
\textbf{ImageNet-C}
\begin{lstlisting}
You are given an image, please generate a short description of the image.
\end{lstlisting}

\textbf{Galaxy Zoo}
\begin{lstlisting}
You are an expert astronomer specializing in galaxy morphology. Please generate a concise description of the image that describes its key visual characteristics.
\end{lstlisting}

\newpage
\section{Mapping between CIFAR-10 and ImageNet classes}\label{appendix:cifar_imgnet_mapping}

In the main text, we extensively consider the setting where we pick images from ImageNet (or its corrupted version) that overlap with CIFAR-10 categories and prompt VLMs to classify the images into CIFAR-10 categories. The exact mapping is presented in Table.~\ref{table:class_mapping}. Notice that here we ignored the deer and horse categories from CIFAR-10 as we cannot find exact mapping in ImageNet categories.

\begin{table*}[h]
\centering
\small
\renewcommand{\arraystretch}{1.5}
\caption{Mapping from CIFAR-10 categories to corresponding ImageNet classes (ImageNet class indices in parentheses).}
\begin{tabular}{ll}
\toprule
\textbf{CIFAR-10 Category} & \textbf{ImageNet Classes (Indices)} \\
\midrule
Airplane & Airliner (404), Warplane (895) \\
Automobile & Beach Wagon (436), Convertible (511), Model T (661), Sports Car (817) \\
Bird & Jay (10), Magpie (11), Eagle (12), Vulture (13), \\
     & Additional Birds (92, 93, 94, 95, 96) \\
Cat & Tabby Cat (281), Tiger Cat (283), Persian Cat (284), Siamese Cat (285) \\
Dog & Chihuahua (151), [Every 20th dog class up to] Mexican Hairless (268) \\
Frog & Bullfrog (30), Tree Frog (31) \\
Ship & Container Ship (510), Liner (628), Pirate Ship (724), \\
     & Schooner (780), Submarine (833) \\
Truck & Fire Truck (555), Garbage Truck (569), Moving Van (675), Pickup Truck (717), \\
      & Police Van (734), Tow Truck (864), Trailer Truck (867) \\
\bottomrule
\end{tabular}
\label{table:class_mapping}
\end{table*}

\newpage
\section{Anomaly ECG inputs construction}\label{appendix:anomaly_construction}
To construct anomaly inputs that are clearly not ECG signal, we use a two step routine: We begin by randomly choose a random number generator
\begin{verbatim}
def get_random_number(size=128):
    rng_list = [
        np.random.normal, np.random.gamma, np.random.exponential,
        np.random.poisson, np.random.uniform, np.random.chisquare,
        np.random.geometric
    ]
    rng = np.random.choice(rng_list)
    return rng(0.6, size=size)
\end{verbatim}

Then we generate a random line figure, with 128 points, as the anomaly inputs via
\begin{verbatim}
plt.plot(get_random_number(128), get_random_number(128), linewidth=0.8, c='black')
\end{verbatim}

\newpage
\section{Galaxy Zoo explanation}\label{appendix:gz2_explain}
Galaxy Zoo is a citizen science dataset, for each galaxy image, annotators are asked to follow the decision tree (\url{https://data.galaxyzoo.org/gz_trees/gz_trees.html}) and perform classification for the galaxy image based on the visual features. Importantly, the metadata of the dataset provides a property called \texttt{leaf\_prob} that describes the agreement level among annotators for the mostly agreed category. We later use \texttt{leaf\_prob} as a score that denotes the inherent ambiguity level of a given galaxy image. 

To actually construct the dataset for evaluation VLM, we first select all images whose majority vote label falls into one of the following categories:
\begin{verbatim}
smooth_round, smooth_cigar, edge_on_disk, unbarred_spiral
\end{verbatim}
which are the four representative galaxy categories studied in Galaxy MNIST~\citep{galaxy_mnist}. Then we \emph{randomly} selected 5,000 images from the pool.

Then, based on the \texttt{leaf\_prob}, we create three levels of disagree level
\begin{verbatim}
Leaf prob in (0.75, 1.0): Disagree level 1
Leaf prob in (0.5, 0.75): Disagree level 2
Leaf prob in (0.0, 0.5) : Disagree level 3
\end{verbatim}

The distribution of categories and \texttt{leaf\_prob} from the 5,000 samples are plotted in Fig.~\ref{fig:gz2_cap}

\begin{figure}[h]
    \centering
    \includegraphics[width=\linewidth]{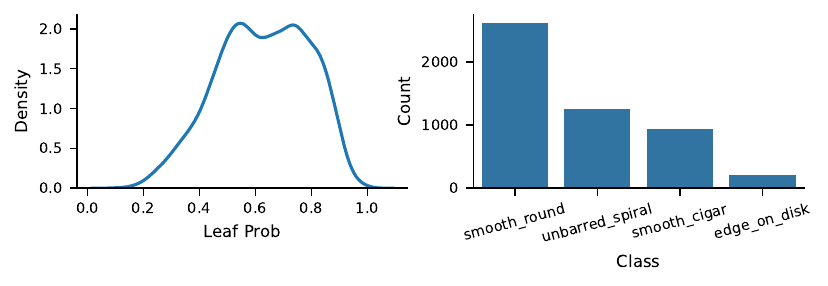}
    \caption{Statistics of the 5,000 sample galaxy zoo datasets used in the experiments.}
    \label{fig:gz2_stat}
\end{figure}

\newpage

\section{Ablation study: Effect of instruction prompt}\label{appendix:abl_study}

\begin{figure*}[h]
    \centering
    \begin{subfigure}{\linewidth}
        % \quad\quad\quad\quad\quad\quad\quad\quad\quad\quad
        \quad\quad
        \includegraphics[height=1.2cm]{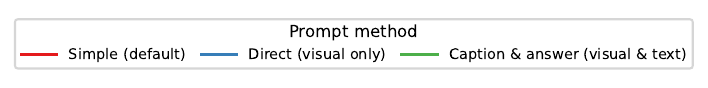}\medskip
        \includegraphics[height=1.2cm]{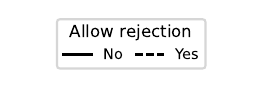}
    \end{subfigure}\\[-2.0ex]
    \centering
    \includegraphics[width=\linewidth]{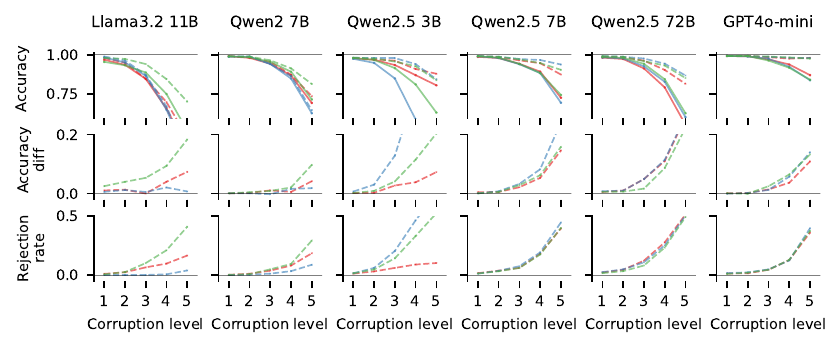}
    \\[-1.5ex]
    \caption{
    \textbf{Ablation study on the prompting strategy.} Similar to the setting studied in Fig.~\ref{fig:corrputed_acc}, here we consider different prompting strategies (line color, detailed in Appendix.~\ref{appendix:abl_study}), and models (columns). The second row illustrates the accuracy improvement, we can see that for certain old models, the prompting strategy matters, but for other, especially the newer ones, it matters little.
    }
    \label{fig:prompt_ablation}
\end{figure*}

\begin{table*}[h]
\small
\centering
\begin{tabular}{lcccccc}
\toprule
\multirow{2}{*}{Model} &
  \multicolumn{2}{c}{Simple} &
  \multicolumn{2}{c}{Direct} &
  \multicolumn{2}{c}{Caption \& Answer} \\
& \textbf{Precision} $\uparrow$ & \textbf{Recall} $\uparrow$
& \textbf{Precision} $\uparrow$ & \textbf{Recall} $\uparrow$
& \textbf{Precision} $\uparrow$ & \textbf{Recall} $\uparrow$
\\
\hline
Llama 3.2 & 0.991 & 0.718 & 0.994 & 0.492 & 0.994 & 0.897 \\
Qwen 2 & 0.964 & 0.757 & 0.994 & 0.215 & 0.992 & 0.917 \\
Qwen 2.5 3B &  0.993 & 0.782 & 0.994 & 0.549 & 0.993 & 0.725 \\
Qwen 2.5 7B & 0.982 & 0.967 & 0.992 & 0.837   & 0.992 & 0.972 \\
Qwen 2.5 72B & 0.976 & 0.986 & 0.969 & 0.992 & 0.987 & 0.987 \\
GPT4o-mini & 0.964 & 0.974 & 0.974 & 0.975 & 0.988 & 0.974 \\
\end{tabular}
\caption{
Similar to the setting in Table.~\ref{tab:ood_detection}, but here we considered different prompting strategy, as discussed in Appendix.~\ref{appendix:abl_study}. Overall, smaller and older models such as Llama 3.2 and Qwen 2 are sensitive to the prompting strategy, more powerful models are less sensitive (e.g.\ GPT4o-mini and Qwen 2.5 72B).
% we prompt different VLMs to differentiate between in v.s. out-of-distribution inputs by giving ``unknown'' as the classification results of OOD samples. Prompting the model to \emph{first caption the image, then answer the query based on the caption} (second row) gives performance better than asking the model to \emph{directly answer the query} (first row). Notice that GPT4 shows a smaller margin between the two methods, indicating that GPT could be implicitly reasoning using the caption.
}
\label{tab:ood_detection_abl}
\end{table*}  
  
We perform an ablation study where we vary the prompt that instructs the VLM to answer the query in a certain way. To be more specific, we considered the following three regimes:
\begin{itemize}[leftmargin=*, topsep=0pt, parsep=0pt, itemsep=1.5pt]
    \item \textbf{Simple} Simple and standard way: we prompt the model to provide step-by-step reasoning~\citep{wei2022chain} and then provide a classification answer.
    \item \textbf{Direct} We prompt the model to \emph{only output the classification answer}, which prohibits the model from generating any explicit intermediate verbal reasoning steps, including image caption.
    \item \textbf{Caption \& answer} We explicitly prompt the model to \emph{always first caption the image}, then answer the question using both the caption and the image.
\end{itemize}

The goal is to understand whether:
\begin{itemize}[leftmargin=*, topsep=0pt, parsep=0pt, itemsep=1.5pt]
    \item The model can perform the uncertainty quantification task through only reading the image (Direct prompting)
    \item The model cannot understand the uncertainty by just reading the image, but would need to first caption the image and then utilize the textual description to determine the level of ambiguity. 
\end{itemize}

In Fig.~\ref{fig:prompt_ablation}, we present accuracy with and without the rejection option (solid and dashed lines) in the first row v.s. corruption level, the improvement of allowing the rejection option (second row) and the rejection rate (third row) under different models (columns) and prompts (line colors). Particularly, for Llama 3.2 and Qwen2, the direct prompt (blue lines) shows almost no improvement when rejection option is enabled and a low rejection rate, and the models only start to reject ambiguous inputs under \textbf{Caption \& answer} (green line), whereas for other models, the VLMs demonstrate the ability of rejection under all types of prompts. This implies that these two models (Llama 3.2 and Qwen2) lack the ability to ``silently'' reason about uncertainty, instead, they need to to first (be asked to) spell the image out, then use the textual feature to determine uncertainty. 
The observation is again confirmed in the OOD detection task (Table.~\ref{tab:ood_detection_abl}), the metrics under the direct prompt for these two models are outperformed by the Caption \& answer prompt.

Regardless, the ability of rejection from the latest models does not seem to be affected by the prompting strategy. It is also worth noting that, the fact that these models can ``silently'' quantify uncertainty does not rule out the possibility that they are verbal thinker internally: They could be ``implicitly'' using verbal reasoning inside.

\newpage
\section{Full results across all corruptions}\label{appendix:all_corruption}
\begin{figure*}[h]
    \centering
    \begin{subfigure}{.95\linewidth}
    \centering
        \includegraphics[width=.25\linewidth]{figs/prompt_legend_new.pdf}
    \end{subfigure}\\[-1.8ex]
    \centering
    % \centering
    % \begin{subfigure}{.8\linewidth}
    %     \includegraphics[width=\linewidth]{figs/ensemble_vs_metric_compact_gaussian_noise.pdf}
    %     \caption{Gaussian noise}
    % \end{subfigure}
    \begin{subfigure}{.8\linewidth}
        \includegraphics[width=\linewidth]{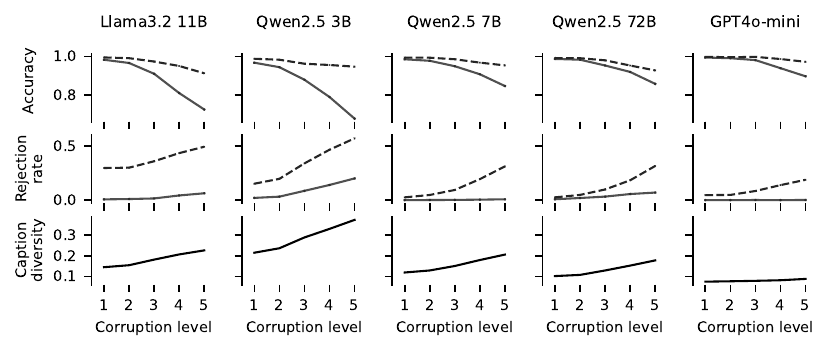}
        \caption{Defocus blur}
    \end{subfigure}
    \begin{subfigure}{.8\linewidth}
        \includegraphics[width=\linewidth]{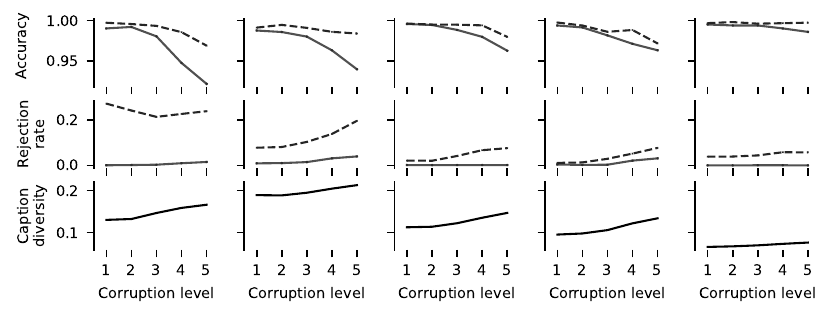}
        \caption{Pixelate}
    \end{subfigure}
    \begin{subfigure}{.8\linewidth}
        \includegraphics[width=\linewidth]{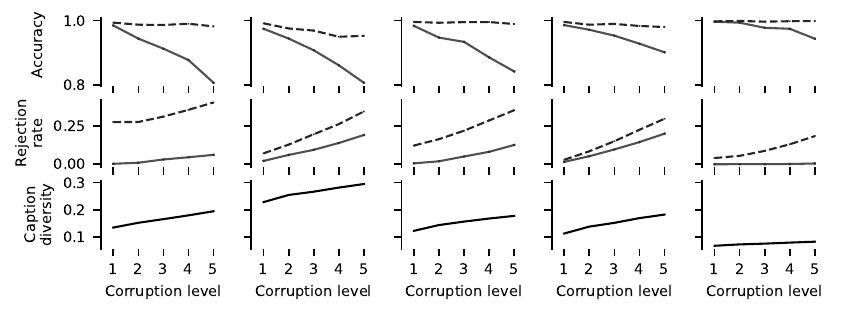}
        \caption{Pixmix}
    \end{subfigure}
    \caption{
    Similar to the setting in Fig.~\ref{fig:corrputed_acc}, but under different corruption types. The observation still holds: Enabling rejection allows VLM to abstain ambiguous samples, providing nearly perfect accuracy for classified samples. Additionally, caption diversity, as we predicted in Sec.~\ref{sec:caption_diversity}, steadily increases as corruption level intensifies.
    }
    \label{fig:imgnet_c_results}
\end{figure*}

\newpage
\section{Licenses for existing assets}\label{appendix:lisce}

\textbf{Datasets:}
\begin{itemize}
    \item \textbf{ImageNet / ImageNet-C} \\
    License: Custom ImageNet Terms of Use (Non-commercial research only) \\
    URL: \url{https://www.image-net.org/download}
    
    \item \textbf{CIFAR-10} \\
    License: MIT License \\
    URL: \url{https://www.cs.toronto.edu/~kriz/cifar.html}
    
    \item \textbf{Galaxy Zoo / Galaxy MNIST} \\
    License: Creative Commons Attribution-NonCommercial-ShareAlike (CC BY-NC-SA 4.0) \\
    URL: \url{https://data.galaxyzoo.org/}
    
    \item \textbf{PTB-XL (ECG Dataset)} \\
    License: PhysioNet Credentialed Health Data License (Restricted; requires credentialed access) \\
    URL: \url{https://physionet.org/content/ptb-xl/}
\end{itemize}

\textbf{Models:}
\begin{itemize}
    \item \textbf{GPT-4o-mini} \\
    License: Proprietary (OpenAI Terms of Use) \\
    URL: \url{https://openai.com}
    
    \item \textbf{Llama 3.2 (Meta)} \\
    License: Meta Llama 3 Community License (Non-commercial research use only) \\
    URL: \url{https://ai.facebook.com/resources/models-and-libraries/llama}
    
    \item \textbf{Qwen 2 / Qwen 2.5 (Alibaba)} \\
    License: Qwen License Agreement (Permits research and certain commercial uses) \\
    URL: \url{https://qwen.aliyun.com/}
\end{itemize}

\end{document}